# A Fast Synchronization Clustering Algorithm

Xinquan Chen[1, 2]

[1]Web Science Center, University of Electronic Science & Technoloughy of China, China

[2]School of Computer Science & Engineering, Chongqing Three Gorges University, China

chenxqscut@126.com

**Abstract**: This paper presents a Fast Synchronization Clustering algorithm (FSynC), which is an improved version of SynC algorithm. In order to decrease the time complexity of the original SynC algorithm, we combine grid cell partitioning method and Red-Black tree to construct the near neighbor point set of every point. By simulated experiments of some artificial data sets and several real data sets, we observe that FSynC algorithm can often get less time than SynC algorithm for many kinds of data sets. At last, it gives some research expectations to popularize this algorithm.

## 1. Introduction

Clustering is an unsupervised learning method that tries to find some obvious distributions and patterns in unlabeled data sets by maximizing the similarity of the objects in a common cluster and minimizing the similarity of the objects in different clusters.

The traditional clustering algorithms are usually categorized into partitioning methods, hierarchical methods, density-based methods, grid-based methods, and model-based methods.

This paper researchs an improved technique of SynC algorithm, which is a famous synchronization clustering algorithm presented in [1]. The major contributions of this paper can be summarized as follows:

(1) It presents a Fast Synchronization Clustering algorithm (FSynC), which is an improved version of SynC algorithm, by combining grid cell partitioning method and Red-Black tree in the process of constructing the near neighbor point set of every point.

(2) It presents another cluster order parameter (named as *t* step average length of edges) to characterize the degree of local synchronization.

(3) It validates the improved effect of FSynC algorithm in time cost by the

simulated experiments of several different kinds of data sets.

The remainder of this paper is organized as follows. Section 2 lists some related papers. Section 3 gives some basic concepts. Section 4 introduces FSynC algorithm. Section 5 validates FSynC algorithm by some simulated experiments. Conclusions and future works are presented in Section 6.

## 2. Related Work

This paper is built on two papers [1, 2]. Chen [2] presents an efficient graph-based clustering method by using the idea of near neighbors and the principle that the global distribution can be approximately represented by many conjoint local distributions. In [2], an efficient method is presented to construct $\delta$ near neighbor point set by using some index structures and effective methods.

Böhm et al. [1] presents a novel clustering approach, SynC algorithm [1], inspired the synchronization principle. SynC algorithm can find the intrinsic structure of the data set without any distribution assumptions and handle outliers by dynamic synchronization. In order to implement automatic clustering, those natural clusters and parameters setting can be discovered by using the Minimum Description Length principle (MDL) [3].

## 3. Some Basic Concepts

Suppose there is a data set $S = \{X_1, X_2, \ldots, X_n\}$ in an $d$-dimensional Euclidean space. Naturally, we can use Euclidean metric as our dissimilarity measure. In order to describe our algorithm clearly, some concepts are presented first.

**Definition 1** [2]. The $\delta$ near neighbor point set $\delta(P)$ of point $P$ is defined as

$$\delta(P) = \{X \mid 0 < dis(X, P) \leq \delta, X \in S\}, \qquad (1)$$

where $dis(X, P)$ is the dissimilarity measure between point $X$ and point $P$ in the data set $S$. $\delta$ is a predefined threshold parameter.

**Definition 2** [1]. The extensive Kuramoto model for clustering is defined as

Point $X = (x_1, x_2, \ldots, x_d)$ is a vector in $d$-dimensional Euclidean space. If each point $X$ is regarded as a phase oscillator according to Eq.(1) of [1], with an interaction in $\delta$ near neighbor point set $\delta(X)$, then the dynamics of the $k$ dimension $x_k$ ($k = 1, 2, \ldots, d$) of the point $X$ over time is described by:

$$x_k(t+1) = x_k(t) + \frac{1}{|\delta(X(t))|} \sum_{Y \in \delta(X(t))} \sin(y_k(t) - x_k(t)), \qquad (2)$$

where $X(t = 0) = (x_1(0), x_2(0), \ldots, x_d(0))$ represents the original phase of point $X$, and $x_k(t+1)$ describes the renewal phase value in $k$-th dimension of point $X$ at the $t$ step evolution.

**Definition 3**. The $t$-step $\delta$ near neighbor undirected graph $G_\delta(t)$ of the data set $S = \{X_1, X_2, \ldots, X_n\}$ is defined as

$$G_\delta(t) = (V(t), E(t)), \qquad (3)$$

where $V(t = 0) = S = \{X_1, X_2, \ldots, X_n\}$ is the original vertex set, $E(t = 0) = \{(X_i, X_j) \mid X_j \in \delta(X_i), X_i \ (i = 1, 2, \ldots, n) \in V\}$ is the original edge set. $V(t) = \{X_1(t), X_2(t), \ldots, X_n(t)\}$ is the $t$-step vertex set of $S$, $E(t) = \{(X_i(t), X_j(t)) \mid X_j(t) \in \delta(X_i(t)), X_i(t) \ (i = 1, 2, \ldots, n) \in V(t)\}$ is the $t$-step edge set, and the weight-computing equation of edge $(X_i, X_j)$ is $weight(X_i, X_j) = dis(X_i, X_j)$.

**Definition 4**. The $t$ step average length of edges, $AveLen(t)$, in $\delta$ near neighbor undirected graph $G_\delta(t)$ is defined as

$$AveLen(t) = \frac{1}{|E(t)|} \sum_{e \in E(t)} |e|, \qquad (4)$$

where $E(t)$ is the $t$ step edge set, and $|e|$ is the length (or weight) of edge $e$. The average length of edges in $G_\delta(t)$ will decreases to its limit zero, $AveLen(t) \to 0$, as more $\delta$ near neighbor points synchronize together with time evolution. This concept is developed independently, although we find that it is equivalent with the cluster order parameter [1] later.

**Definition 5** [1]. The cluster order parameter $r_c$ characterizing the degree of local synchronization is defined as:

$$r_c = \frac{1}{n} \sum_{i=1}^{n} \sum_{Y \in \delta(X)} e^{-dis(X,Y)}, \qquad (5)$$

The value of $r_c$ increases to its limit 1 as more $\delta$ near neighbor points synchronize together with time evolution.

**Definition 6** [2]. The Grid Cell is defined as follows:

Grid cells can be obtained after partitioned the multidimensional ordered-attribute space by using a multidimensional grid partition method.

The data structure of grid cell $g$ can be defined as:

$$DS(g) = (Grid\_Label, Grid\_Position, Grid\_Range, Point\_Number, Points\_Set). \qquad (6)$$

In equation (6),

Grid_Label is the key label of the grid cell.

Grid_Position is the center position of the grid cell. It is a $d$-dimensional vector expressed by $P = (p_1, p_2, …, p_d)$.

Grid_Range records the region of the grid cell. It is a $d$-dimensional interval vector expressed by:

$$R = ([p_1-r_1/2, p_1+r_1/2), … , [p_d-r_d/2, p_d+r_d/2)), \qquad (7)$$

where $r_i$ ($i =1, 2, …, d$) is the interval length in the $i$-th dimension of the grid cell.

Point_Number records the number of points of the grid cell.

Points_Set records the labels of points of the grid cell. In FSynC algorithm, we use a Red-Black tree to records the labels of points of the grid cell to obtain efficient inserting and deleting operations.

## 4. A Fast Version of Synchronization Clustering Algorithm

In order to implement a fast version of SynC algorithm, first we partition the data space of the data set $S = \{X_1, X_2, …, X_n\}$ by using a kind of grid partitioning method. Then construct an effective index of all grid cells and compute $\delta$ near neighbor grid cell set of each grid cell. In each synchronization step of dynamical clustering, there are insert and delete operations of some data points. These data points disengage their original grid cells and enter new grid cells because of their synchronization moving. If each grid cell uses a Red-Black tree to store its data points in each synchronization step, then constructing the near neighbor point set of every point will became more quick when the number of data points is enough large.

Although we use the Euclidean metric as our dissimilarity measure in this paper, the algorithm is by no means restricted to this measure and this data space. If we can construct a proper dissimilarity measure in a hybrid-attribute space, then the algorithm can also be used.

**4.1 The Description of FSynC Algorithm**

**Algorithm Name**: Fast Synchronization Clustering algorithm (FSynC)

**Input**: data set $S = \{X_1, X_2, …, X_n\}$, interval length vector of grid cell $Interval = (r_1, r_2, …, r_d)$, dissimilarity measure $d(·, ·)$, parameters $\delta$.

**Output**: The final convergent result $S(T) = \{X_1(T), X_2(T), …, X_n(T)\}$ of the original data set $S$.

**Procedure**:

Step1. Partition the data space of the data set $S = \{X_1, X_2, …, X_n\}$ by using a multidimensional grid partitioning method basing on the interval length vector of grid cell *Interval* = ($r_1$, $r_2$, …, $r_d$). Suppose we obtain *N* grid cells. Usually, multidimensional index tree or multidimensional array can be used to as an index of the *N* grid cells.

Step2. Constructing $\delta$ near neighbor grid cell set for each grid cell. The *N* $\delta$ near neighbor grid cell sets can be used to construct *n* $\delta$ near neighbor point sets with less time cost in the next repeat synchronization clustering procedure. The details of Step1 and Step2 are described in [2].

Step3. When the dynamical clustering does not reach its convergent result, repeat the synchronization clustering procedure listed in Figure 1.

```
1    IterateStep is set as zero firstly: t ← 0;
2    for (i = 1; i ≤ n; i++)
3        X_i(t) ← X_i;
4    while (the dynamical clustering does not reach its convergent result)
5    {
6        for (i = 1; i ≤ n; i++)
7        {
8            Locate the corresponding grid cell for X_i(t);
9            X_i(t) is inserted into the Red-Black tree of its grid cell;
10       }
11       Construct the δ near neighbor point set δ(X_i(t)) for each point X_i(t) (i = 1, 2, …, n)
    according to Definition 1 and the δ near neighbor grid cell set of the grid cell of X_i(t);
12       for (i = 1; i ≤ n; i++)
13           Compute the new value, X_i(t+1), of X_i(t) using Eq.(2);
14       for (i = 1; i ≤ n; i++)
15           if (X_i(t+1) disengages the original grid cell of X_i(t) and enter a new grid cell)
16           {
17               X_i(t) is deleted from the Red-Black tree of its original grid cell;
18               The number of points of the original grid cell decreases with 1;
19               X_i(t+1) is inserted into the Red-Black tree of its new grid cell;
20               The number of points of the new grid cell increases with 1;
21           }
22       Compute the t step average length of edges of all points, AveLen(t), using Eq.(4);
         /* We can also compute the cluster order parameter r_c according to Definition 5 [2]
    instead of computing AveLen(t). */
23       IterateStep is increased with one: t++;
24       if (AveLen(t) → 0)   /* AveLen(t) → 0 is equivalent with r_c → 1. */
25           We think the dynamical clustering is reaching its convergent result;
26   }
```

Figure 1. The repeat synchronization clustering procedure in FSynC algorithm

Step4. Finally we get a convergent result $S(T) = \{X_1(T), X_2(T), …, X_n(T)\}$, where *T* is the times of the while circulation in Figure 1. The final convergent set $S(T)$ reflects the natural clusters or isolate points of the data set *S*.

**4.2 Some basic knowledge of FSynC Algorithm**

**Lemma 1**. Suppose a function $f(x_1, x_2, …, x_m) = x_1^{x_1} \cdot x_2^{x_2} \cdot … \cdot x_m^{x_m}$, subject to $x_1$

$+ x_2 + \ldots + x_m = n$. Then its maximum is $(n/m)^n$ when $x_j = (n/m), j = 1, 2, \ldots, m$.

**Proof**: According to Lagrange method, we can prove this lemma easily. We first set

$$L(x_1, x_2, \ldots, x_m; \lambda) = x_1^{x_1} \cdot x_2^{x_2} \cdot \ldots \cdot x_m^{x_m} + \lambda * (x_1 + x_2 + \ldots + x_m - n).$$

Because

$$\frac{\partial L}{x_1} = (1 + \ln x_1) \cdot x_1^{x_1} \cdot x_2^{x_2} \cdot \ldots \cdot x_m^{x_m} + \lambda = 0,$$

$$\frac{\partial L}{x_2} = (1 + \ln x_2) \cdot x_1^{x_1} \cdot x_2^{x_2} \cdot \ldots \cdot x_m^{x_m} + \lambda = 0,$$

……

$$\frac{\partial L}{x_m} = (1 + \ln x_m) \cdot x_1^{x_1} \cdot x_2^{x_2} \cdot \ldots \cdot x_m^{x_m} + \lambda = 0,$$

$$x_1 + x_2 + \ldots + x_m = n.$$

So there are

$$(1 + \ln x_1) = (1 + \ln x_2) = \ldots = (1 + \ln x_m),$$

and $x_1 + x_2 + \ldots + x_m = n$.

At last, we get

$$x_j = (n/m), j = 1, 2, \ldots, m.$$

At this time, the maximum is $(n/m)^n$.

**Theorem 1**. Suppose there are $N$ grid cells after partitioned the $d$-dimensional ordered-attribute space by using a multidimensional grid partition method. If the data set $S = \{X_1, X_2, \ldots, X_n\}$ is indexed initially by $m$ ($m \leq N$) grid cells and $m$ corresponding Red-Black trees, then constructing the initial $m$ Red-Black trees needs Time = $O(n * log(n) + m)$ and Space = $O(n + m)$.

**Proof**: At first, the data set $S = \{X_1, X_2, \ldots, X_n\}$ is located into $m$ grid cells, which needs Time = $O(n * d)$. Suppose $n_i$ ($n_i > 0$) is the number of points in the $i$-th grid cell. In initial step of clustering, if $n_i$ ia larger than 1, all points in the $i$-th grid cell are inserted into the $i$-th Red-Black tree in turn, which needs Time = $O(log(n_i!)) < O(n_i * log(n_i))$. If $n_i$ is equal to 1, the point in the $i$-th grid cell is inserted into the $i$-th Red-Black tree, which needs Time = $O(1)$. Suppose the number of data points in the first $m^*$ ($m^* \leq m$) grid cells is larger than 1, and the number of data points in the last $m - m^*$ grid cells is equal to 1. For the $m$ grid cells where the number of data points is larger than 0, because

$$n_1 * log(n_1) + n_2 * log(n_2) + \ldots + n_{m^*} * log(n_{m^*}) \leq (n - m + m^*) * log((n - m + m^*) / m^*) < n * log(n / m^*) < n * log(n),$$

and

$$n_{(m^*+1)} + n_{(m^*+2)} + \ldots + n_m = m - m^* < m.$$

So constructing the initial $m$ Red-Black trees for the data set $S = \{X_1, X_2, \ldots, X_n\}$ needs Time = $O(n * log(n) + m)$ and Space = $O(n + m)$.

**Theorem 2**. Suppose there are $N$ grid cells after partitioned the $d$-dimensional ordered-attribute space by using a multidimensional grid partition method. If the data set $S = \{X_1, X_2, \ldots, X_n\}$ is indexed initially by $m$ ($m \leq N$) grid cells and $m$ corresponding Red-Black trees, then the inserting and deleting operations in dynamical clustering need Time = $O(n * log(n / m))$ and Space = $O(n + N)$.

**Proof**: In dynamical clustering, usually, there are only part points will be deleted from their original grid cells and be inserted into new grid cells. A special case is that all points in the $i$-th grid cell are deleted from the $i$-th Red-Black tree and are inserted into the $j$-th Red-Black tree, which needs Time = $O(n_i * log(n_i) + n_i * log(n_i + n_j)) < O(n_i * log(n))$. So the inserting and deleting operations in dynamical clustering need Time = $O(n * log(n))$ and Space = $O(n + N)$.

**4.3 Time complexity analysis of FSynC Algorithm**

According to [1] and our analysis, the original SynC Algorithm [1] needs Time = $O(Td\, n^2)$. Our FSynC algorithm uses a strategy of "space exchanges time".

In Step1, according to [2], we know that partitioning the data space and storing the basic information of all grid cells according to Definition 6 need Time = $O(nd + Nd)$ and Space = $O(nd + Nd)$. According to Theorem 1, assigning all data points to their corresponding grid cells needs Time = $O(nd + nlog(n) + m)$ and Space = $O(nd + m)$.

In Step2, if we use a simple method, then constructing $\delta$ near neighbor grid cell sets for every grid cell needs Time = $O(dN^2)$ and Space = $O(Nd)$. If we use the coordinates-locating method [2], then constructing $\delta$ near neighbor grid cell sets for every grid cell needs Time = $O(Nd + N * C^d)$ and Space = $O(Nd)$, where $C$ is related to $\delta$ and $Interval = (r_1, r_2, \ldots, r_d)$. If $\delta \leq r_i$ ($i = 1, 2, \ldots, d$), then $C = 3$. And if $r_i < \delta \leq 2r_i$ ($i = 1, 2, \ldots, d$), then $C = 5$.

In Step3, locating the corresponding grid cell for $X_i(t)$ needs Time = $O(d)$ [2]. It needs Time = $O(log(\text{number of points in the grid cell of } X_i(t)))$ that $X_i(t)$ is inserted into the Red-Black tree of its grid cell. It also needs Time = $O(log(\text{number of points in}$ the grid cell of $X_i(t)$)) that $X_i(t)$ is deleted from the Red-Black tree of its original grid cell. According to Theorem 2, we know that Step3 needs Time = $O(Tn * (d + log(n / m)))$ and Space = $O(n + N)$, where $T$ is the times of the while circulation in Figure 1 and $m$ is the number of grid cells where the number of data points is larger than 0.

Step4 needs Time = $O(n)$ and Space = $O(n)$.

**4.4 Setting parameters in FSynC Algorithm**

Parameter $\delta$ will affect the results of clusters. In [1], parameter $\delta$ is optimized by the MDL principle [3]. In [4], two other methods can also be used to estimate parameter $\delta$.

Parameter interval length vector of grid cell $Interval = (r_1, r_2, …, r_d)$ will affect the time cost of FSynC algorithm. We know that the whole time cost of FSynC algorithm is Time = $O(nd + nlog(n) + Nd + \min\{dN^2, Nd + N * C^d\} + Tn * (d + log(n / m)))$. Here, $N$ is determined by $Interval = (r_1, r_2, …, r_d)$. In [2], the relation between $Interval = (r_1, r_2, …, r_d)$ and $N$ is discussed in detail.

## 5. Simulated Experiments

**5.1 Experimental Design**

Our experiments are finished in a personal computer (Capability Parameters: Pentium(R) Dual CPU T3200 2.0GHz, 2G Memory). Experimental programs are developed using Visual C++6.0 under Windows XP.

To verify the improvements in time complexity of this algorithm, there will be some experiments of some artificial data sets, two UCI data sets, and two bmp pictures in the next subsections.

Four kinds of artificial data sets (DS1 – DS4) are produced in a 2-D region [0, 600] × [0, 600] by a program. Four kinds of artificial data sets (DS5 – DS8) are produced in a range [0, 600] in each dimension by a similar program. Table 1 is the description of the eight kinds of artificial data sets.

Table 1. The description of eight kinds of artificial data sets

| Data Sets (DS) | Number of Clusters (NC) | With Noise | Cluster Semidiameter (CS) | Dimension ($d$) |
|---|---|---|---|---|
| DS1 | 5 | yes | 40 | 2 |
| DS2 | 5 | no | 50 | 2 |
| DS3 | 9 | yes | 30 | 2 |
| DS4 | 9 | no | 40 | 2 |
| DS5 | 5 | yes | 40 | 1, 2, … ,8 |
| DS6 | 5 | no | 40 | 1, 2, … ,8 |
| DS7 | 9 | yes | 40 | 1, 2, … ,8 |
| DS8 | 9 | no | 40 | 1, 2, … ,8 |

3D_spatial_network and Ttamilnadu Electricity Board Hourly Readings are two UCI data sets [5] used in our experiments.

Two bmp pictures are obtained from Internet.

In SynC algorithm and FSynC algorithm, the times of synchronization clustering in the while circulation of SynC and FSynC algorithms is set as 50 in our simulated experiments.

Comparative results of the two algorithms are given by four figures (Figure 2 - Figure 5) and two table (Table 2 - Table 3), and performance of algorithms is measured by time cost (second).

In subsection 5.2, FSynC algorithm will be compared with SynC algorithm in time cost using some artificial data sets.

In subsection 5.3, FSynC algorithm will be compared with SynC algorithm in time cost using two UCI data sets.

In subsection 5.4, FSynC algorithm will be compared with SynC algorithm in time cost using two bmp pictures.

Since $\delta$ near neighbor point of point $P$ locates in the grid cell of point $P$ or its near grid cells, so ususlly less time is needed to construct $\delta$ near neighbor point sets of all points if we set $r_i$ ($i = 1, 2, …, d$) $\geq \delta$. The detailed discussion on how to construct grid cells is described in [2].

In the experiment, parameter $r_i$ ($i = 1, 2, …, d$) is the interval length in the $i$-th dimension of grid cell [2], and $\delta$ is the threshold parameter in Definition 1. How to select a proper parameter $\delta$ for SynC algorithm is discussed detailly in [1]. Selecting a proper parameter $\delta$, FSynC algorithm can use the same method as SynC algorithm. In FSynC algorithm, different parameter $r_i$ ($i = 1, 2, …, d$) for different dimensions will result in different number of grid cells and different time cost.

**5.2 Compare with SynC Algorithm Using Some Artificial Data Sets (DS1 – DS8)**

Figure 2 and Figure 3 are the experimental results of four artificial data sets (DS1 – DS4) in time cost between FSynC algorithm and SynC algorithm. Table 2 and Table 3 are the experimental results of four artificial data sets (DS5 – DS8) in time cost between FSynC algorithm and SynC algorithm.

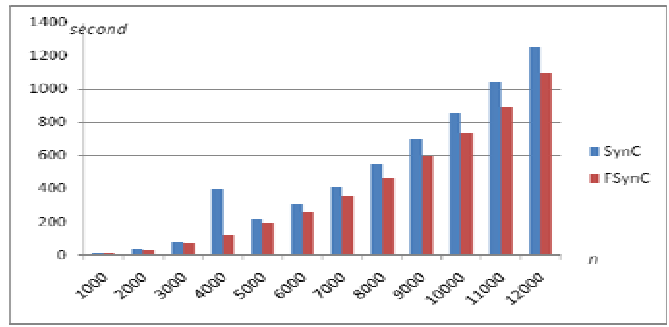

(a). DS1

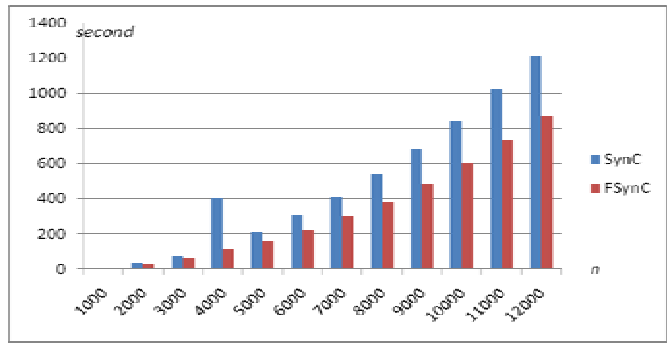

(b). DS2

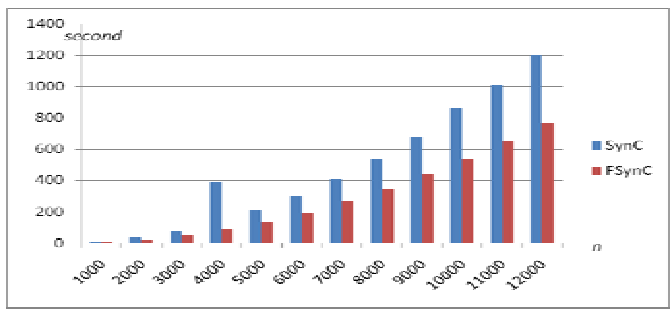

(c). DS3

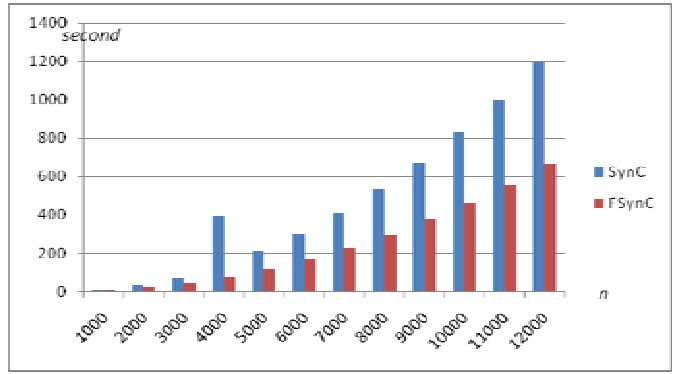

(d). DS4

Figure 2. Comparison of time cost between FSynC algorithm and SynC algorithm by using four kinds of artificial data sets

($\delta = 18$; In FSync, $r_i$ ($i = 1, 2$) = 20)

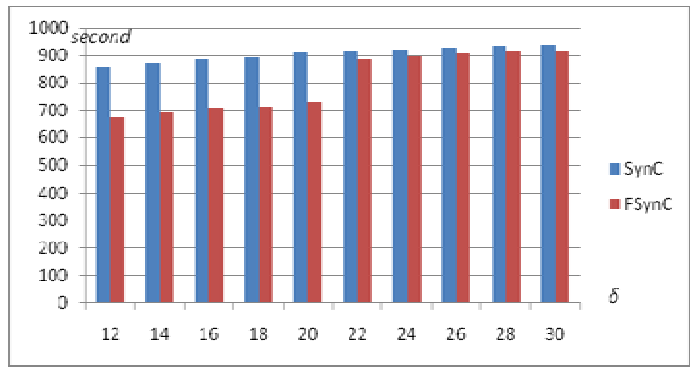

(a). DS1 (In FSync, $N = 648$)

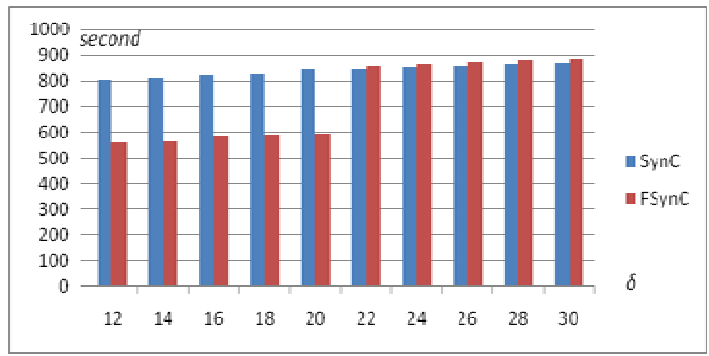

(b). DS2 (In FSync, $N = 420$)

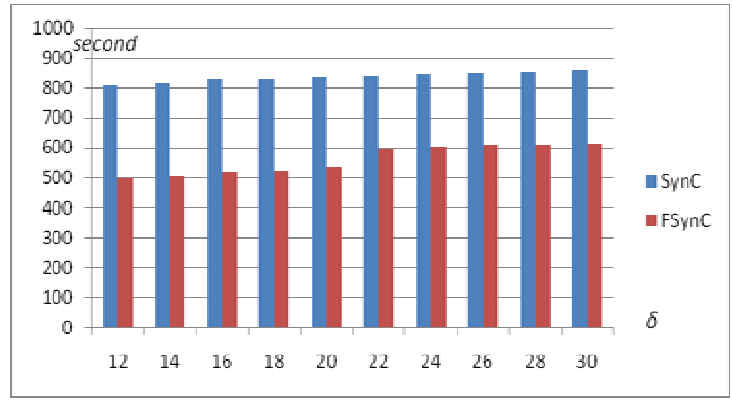

(c). DS3 (In FSync, $N = 812$)

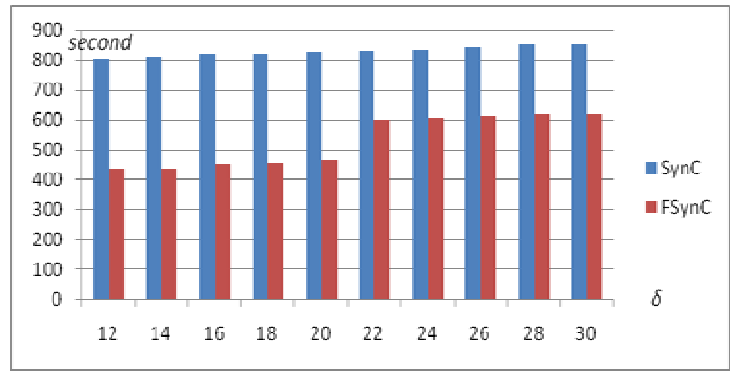

(d). DS4 (In FSync, $N = 783$)

Figure 3. Comparison of time cost between FSynC algorithm and SynC algorithm by using four kinds of artificial data sets

($n = 10000$; In FSync, $r_i$ ($i = 1, 2$) = 20)

Table 2. Comparison of time cost between FSynC algorithm and SynC algorithm by using four kinds of data sets

(a) DS5

| $d$ | $ST$ of DS5 (second) ($n = 6000$; $\delta = 18$) | | | | $ST$ of DS5 (second) ($n = 12000$; $\delta = 18$) | | | |
|---|---|---|---|---|---|---|---|---|
| | $SynC$ | $FSynC$ | | | $SynC$ | $FSynC$ | | |
| | $ST$ | $ST$ | $N$ | $r_i$ | $ST$ | $ST$ | $N$ | $r_i$ |
| 1 | 218 | 236 | 525 | 1 | 882 | 926 | 525 | 1 |
| | | 263 | 175 | 3 | | 1012 | 175 | 3 |
| | | 290 | 53 | 10 | | 1165 | 53 | 10 |
| | | 310 | 27 | 20 | | 1257 | 27 | 20 |
| 2 | **199** | **198** | 2688 | 10 | **800** | **798** | 3240 | 10 |
| | | 218 | 672 | 20 | | 894 | 810 | 20 |
| 3 | **313** | **283** | 6498 | 30 | **1250** | **1113** | 4864 | 30 |
| 4 | **319** | **316** | 15840 | 50 | **1280** | **1142** | 10560 | 50 |
| | | **282** | 1080 | 100 | | **1138** | 720 | 100 |
| 5 | **401** | **298** | 9072 | 100 | **1602** | **1137** | 7560 | 100 |
| 6 | **352** | **324** | 12500 | 120 | **1414** | **1182** | 15625 | 120 |
| | | 397 | 4096 | 150 | | 1528 | 4096 | 150 |
| 7 | **411** | **374** | 16384 | 150 | **1643** | **1309** | 16384 | 150 |
| | | 598 | 2187 | 200 | | 2426 | 2187 | 200 |
| 8 | 449 | 1694 | 65536 | 150 | **1804** | **1724** | 32768 | 150 |
| | | 747 | 6561 | 200 | | 3156 | 4374 | 200 |

(b) DS6

| $d$ | $ST$ of DS6 (second) ($n = 6000$; $\delta = 18$) | | | | $ST$ of DS6 (second) ($n = 12000$; $\delta = 18$) | | | |
|---|---|---|---|---|---|---|---|---|
| | $SynC$ | $FSynC$ | | | $SynC$ | $FSynC$ | | |
| | $ST$ | $ST$ | $N$ | $r_i$ | $ST$ | $ST$ | $N$ | $r_i$ |
| 1 | 221 | 233 | 525 | 1 | 881 | 921 | 525 | 1 |
| | | 263 | 175 | 3 | | 1007 | 175 | 3 |
| | | 289 | 53 | 10 | | 1166 | 53 | 10 |
| | | 311 | 27 | 20 | | 1258 | 27 | 20 |
| 2 | **200** | **184** | 6206 | 5 | **800** | **712** | 6206 | 5 |
| | | **199** | 1566 | 10 | | **800** | 1566 | 10 |
| | | 220 | 405 | 20 | | 891 | 405 | 20 |
| 3 | **313** | **279** | 4320 | 30 | **1254** | **1124** | 4320 | 30 |
| 4 | **325** | **288** | 3168 | 50 | **1332** | **1144** | 3168 | 50 |
| | | **292** | 216 | 100 | | **1149** | 216 | 100 |
| 5 | **401** | **294** | 7560 | 100 | **1605** | **1142** | 7560 | 100 |
| 6 | **352** | 393 | 21600 | 100 | **1417** | **1242** | 21600 | 100 |
| | | **333** | 6000 | 120 | | **1322** | 6000 | 120 |
| | | 398 | 3072 | 150 | | 1580 | 3072 | 150 |
| 7 | **412** | **407** | 19200 | 140 | **1646** | **1327** | 19200 | 140 |
| | | 456 | 12288 | 150 | | 1697 | 12288 | 150 |
| | | 597 | 1458 | 200 | | 2426 | 1458 | 200 |
| 8 | **449** | **417** | 16200 | 130 | **1797** | **1417** | 16200 | 130 |
| | | 495 | 3072 | 150 | | 1946 | 3072 | 150 |

(c) DS7

| $d$ | ST of DS7 (second) ($n = 6000; \delta = 18$) | | | | ST of DS7 (second) ($n = 12000; \delta = 18$) | | | |
|---|---|---|---|---|---|---|---|---|
| | SynC | FSynC | | | SynC | FSynC | | |
| | ST | ST | N | $r_i$ | ST | ST | N | $r_i$ |
| 1 | **212** | **180** | 60 | 10 | **840** | **717** | 58 | 10 |
| | | **197** | 30 | 20 | | **777** | 29 | 20 |
| 2 | **193** | **123** | 3420 | 10 | **771** | **487** | 3480 | 10 |
| | | **140** | 870 | 20 | | **548** | 870 | 20 |
| 3 | **306** | **163** | 8360 | 30 | **1226** | **631** | 8360 | 30 |
| 4 | **316** | **215** | 20592 | 50 | **1282** | **677** | 18876 | 50 |
| | | **213** | 1512 | 100 | | **843** | 1512 | 100 |
| 5 | **399** | **185** | 9072 | 100 | **1594** | **671** | 8820 | 100 |
| 6 | **351** | **226** | 8000 | 150 | **1409** | **746** | 8000 | 150 |
| 7 | **410** | **262** | 6912 | 200 | **1642** | **1004** | 6912 | 200 |
| 8 | **450** | **354** | 11664 | 200 | **1799** | **1338** | 8748 | 200 |
| | | 496 | 6561 | 250 | | 1961 | 6561 | 250 |

(d) DS8

| $d$ | ST of DS8 (second) ($n = 6000; \delta = 18$) | | | | ST of DS8 (second) ($n = 12000; \delta = 18$) | | | |
|---|---|---|---|---|---|---|---|---|
| | SynC | FSynC | | | SynC | FSynC | | |
| | ST | ST | N | $r_i$ | ST | ST | N | $r_i$ |
| 1 | **211** | **181** | 58 | 10 | **850** | **718** | 58 | 10 |
| | | **197** | 29 | 20 | | **779** | 29 | 20 |
| 2 | **193** | **125** | 3078 | 10 | **771** | **482** | 3078 | 10 |
| | | **140** | 783 | 20 | | **559** | 783 | 20 |
| 3 | **307** | **164** | 7128 | 30 | **1226** | **629** | 7128 | 30 |
| 4 | **316** | **165** | 6552 | 50 | **1286** | **737** | 6552 | 50 |
| | | **228** | 504 | 100 | | **935** | 504 | 100 |
| 5 | **398** | **189** | 7560 | 100 | **1592** | **738** | 7560 | 100 |
| 6 | **351** | **217** | 8000 | 150 | **1408** | **814** | 8000 | 150 |
| 7 | **411** | **284** | 3888 | 200 | **1640** | **1107** | 3888 | 200 |
| 8 | **449** | **332** | 6561 | 200 | **1798** | **1283** | 6561 | 200 |

Table 3. Comparison of time cost between FSynC algorithm and SynC algorithm by using four kinds of data sets

(a) DS5 ($d = 1$)

| $\delta$ | ST of DS5 (second) ($n = 6000$) | | | | ST of DS5 (second) ($n = 12000$) | | | |
|---|---|---|---|---|---|---|---|---|
| | SynC | FSynC | | | SynC | FSynC | | |
| | ST | ST | N | $r_i$ | ST | ST | N | $r_i$ |
| 2 | **242** | **58** | 5245 | 0.1 | **968** | **205** | 5245 | 0.1 |
| | | **56** | 525 | 1 | | **206** | 525 | 1 |
| | | 363 | 27 | 20 | | 1438 | 27 | 20 |
| 6 | **248** | **109** | 5245 | 0.1 | **980** | **384** | 5245 | 0.1 |
| | | **133** | 525 | 1 | | **471** | 525 | 1 |
| | | 355 | 27 | 20 | | 1425 | 27 | 20 |
| 10 | **254** | **192** | 5245 | 0.1 | **1015** | **726** | 5245 | 0.1 |
| | | **176** | 525 | 1 | | **692** | 525 | 1 |
| | | 367 | 27 | 20 | | 1458 | 27 | 20 |
| 14 | **259** | 276 | 5245 | 0.1 | **1041** | 1081 | 5245 | 0.1 |
| | | **254** | 525 | 1 | | **1039** | 525 | 1 |
| | | 367 | 27 | 20 | | 1486 | 27 | 20 |
| 18 | **262** | 290 | 5245 | 0.1 | 1053 | 1125 | 5245 | 0.1 |
| | | **208** | 525 | 1 | | 1111 | 525 | 1 |
| | | 369 | 27 | 20 | | 1501 | 27 | 20 |

(b) DS6 ($d = 1$)

| $\delta$ | ST of DS6 (second) ($n = 6000$) | | | | ST of DS6 (second) ($n = 12000$) | | | |
|---|---|---|---|---|---|---|---|---|
| | SynC | FSynC | | | SynC | FSynC | | |
| | ST | ST | N | $r_i$ | ST | ST | N | $r_i$ |
| 2 | **246** | **58** | 5245 | 0.1 | **970** | **203** | 5245 | 0.1 |
| | | **57** | 525 | 1 | | **204** | 525 | 1 |
| | | 365 | 27 | 20 | | 1433 | 27 | 20 |
| 6 | **246** | **105** | 5245 | 0.1 | **977** | **349** | 5245 | 0.1 |
| | | **134** | 525 | 1 | | **477** | 525 | 1 |
| | | 352 | 27 | 20 | | 1405 | 27 | 20 |
| 10 | **256** | **186** | 5245 | 0.1 | **1018** | **728** | 5245 | 0.1 |
| | | **172** | 525 | 1 | | **694** | 525 | 1 |
| | | 367 | 27 | 20 | | 1460 | 27 | 20 |
| 14 | **265** | 274 | 5245 | 0.1 | 1036 | 1085 | 5245 | 0.1 |
| | | **254** | 525 | 1 | | 1039 | 525 | 1 |
| | | 368 | 27 | 20 | | 1489 | 27 | 20 |
| 18 | 265 | 288 | 5245 | 0.1 | 1042 | 1118 | 5245 | 0.1 |
| | | 276 | 525 | 1 | | 1094 | 525 | 1 |
| | | 370 | 27 | 20 | | 1490 | 27 | 20 |

(c) DS6 ($d = 8$)

| $\delta$ | ST of DS6 (second) ($n = 6000$) | | | | ST of DS6 (second) ($n = 12000$) | | | |
|---|---|---|---|---|---|---|---|---|
| | SynC | FSynC | | | SynC | FSynC | | |
| | ST | ST | N | $r_i$ | ST | ST | N | $r_i$ |
| 2 | 531 | 570 | 24300 | 120 | **2126** | **1653** | 24300 | 120 |
| | | 937 | 864 | 200 | | 3828 | 864 | 200 |
| 6 | 530 | 570 | 24300 | 120 | **2131** | **1652** | 24300 | 120 |
| | | 936 | 864 | 200 | | 3828 | 864 | 200 |
| 10 | 531 | 570 | 24300 | 120 | **2126** | **1653** | 24300 | 120 |
| | | 937 | 864 | 200 | | 3821 | 864 | 200 |
| 14 | 530 | 570 | 24300 | 120 | **2124** | **1653** | 24300 | 120 |
| | | 932 | 864 | 200 | | 3821 | 864 | 200 |
| 18 | 531 | 571 | 24300 | 120 | **2134** | **1672** | 24300 | 120 |
| | | 929 | 864 | 200 | | 3820 | 864 | 200 |

## 5.3 Compare with SynC Algorithm Using Two UCI Data Sets

Figure 4 is the experimental results of two UCI data sets in time cost between FSynC algorithm and SynC algorithm.

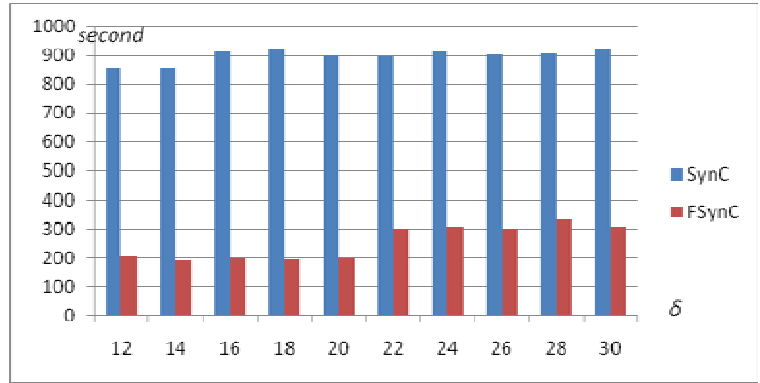

(a). 3D_spatial_network data set ($d = 3$; In FSync, $N = 29791$)

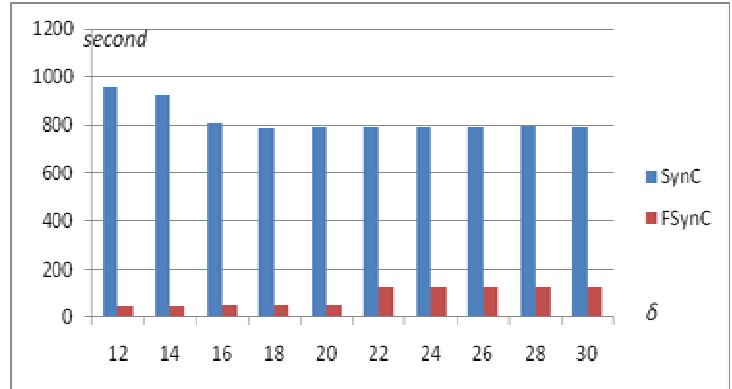

(b). Tamilnadu Electricity Board Hourly Readings data set ($d = 2$; In FSync, $N = 961$)
Figure 4. Comparison of time cost between FSynC algorithm and SynC algorithm by using two UCI data sets (all points are standard into a range [0, 600] in each dimension)
($n = 10000$; In FSync, $r_i$ ($i = 1, 2, …, d$) = 20)

## 5.4 Compare with SynC Algorithm by Clustering Pixel Points of Two Bmp Pictures in RGB Space

Figure 5 is the experimental results of two bmp pictures data sets in time cost between FSynC algorithm and SynC algorithm.

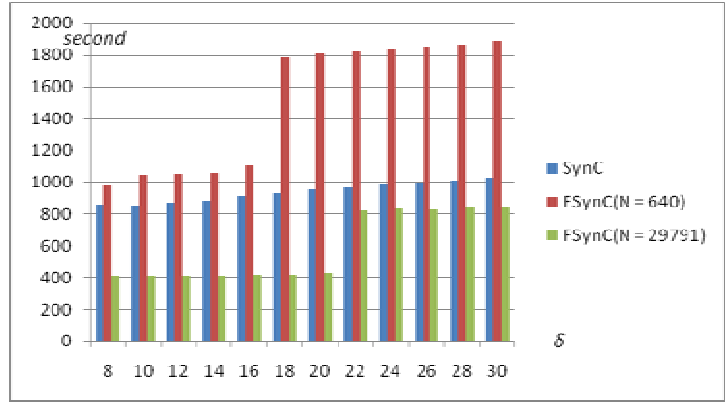

(a). Picture1 data set
(In FSync, $r_i$ ($i$ = 1, 2, 3) = 16, $N$ = 640 (orginal pixel points); $r_i$ ($i$ = 1, 2, 3) = 20 (all pixel points are standard into a range [0, 600] in each dimension), $N$ = 29791)

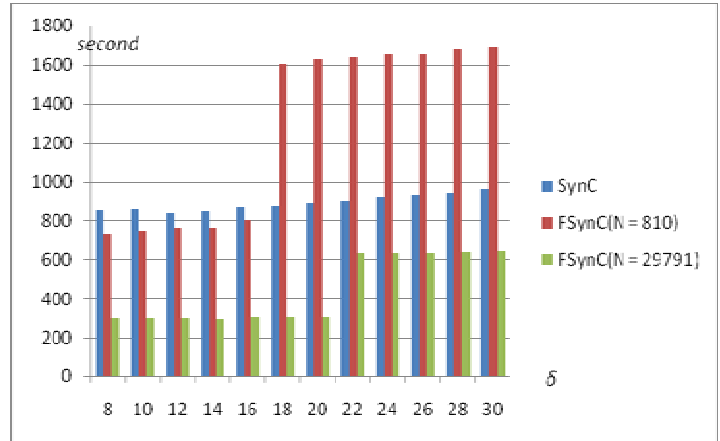

(b). Picture2 data set
(In FSync, $r_i$ ($i$ = 1, 2, 3) = 16, $N$ = 810 (orginal pixel points); $r_i$ ($i$ = 1, 2, 3) = 20 (all pixel points are standard into a range [0, 600] in each dimension), $N$ = 29791)
Figure 5. Comparison of time cost between FSynC algorithm and SynC algorithm by using two picture pixel data sets ($d$ = 3; $n$ = 10000)

## 5.5 Analysis and Conclusions of Experimental Results

From the comparative experimental results of Figure 2, Figure 3, and Table 2, we observe that FSynC algorithm is faster than SynC algorithm for many cases. From the comparative experimental results of Table 3, we find that even for two cases (case1: $d$ = 1 of Table 2 (a); case2: $d$ = 1 of Table 2 (a)), FSynC algorithm can also get a less time cost by selecting another parameter $r_i$ ($i$ = 1, 2, …, $d$) or parameter $\delta$.

From the comparative results of Figure 4, we observe that FSynC algorithm is faster than SynC algorithm for the two UCI data sets.

From the comparative results of Figure 5, we observe that FSynC algorithm can get a less time cost by selecting a proper parameter $r_i$ ($i$ = 1, 2, 3).

FSynC algorithm is an improved clustering algorithm with faster clustering speed than SynC algorithm for many cases. The time cost of FSynC algorithm is sensitive to parameter $r_i$ ($i$ = 1, 2, …, $d$). Usually, if the data sets have obvious clusters, the number of grids is better near to or larger than the number of points. If the number of grids is too less, then perhaps FSynC algorithm can not obtain obvious improvement in time cost.

## 6. Conclusions

This paper presents an improved clustering algorithm, FSynC, which gets the same clustering results and can often obtain faster clustering speed for some kinds of data sets than SynC algorithm.

FSynC algorithm is also robust to outliers and can find obvious clusters with different shapes. The number of clusters does not have to be fixed before clustering. Usually, parameter $\delta$ has some valid interval that can be determined by using an exploring method listed in [4] or using the same method presented in [2]. In the process of constructing $\delta$ near neighbor point sets, the time cost of FSynC algorithm can often be decreased by combining grid cell partitioning method and Red-Black tree index structure.

The next work is to explore the relation between the time cost and the number of grids of this algorithm.